\documentclass[conference]{IEEEtran}
\IEEEoverridecommandlockouts
\usepackage{cite}
\usepackage{amsmath,amssymb,amsfonts}
\usepackage{booktabs}
\usepackage{algorithmic}
\usepackage{graphicx}
\usepackage{textcomp}
\usepackage{pifont}  
\usepackage{xcolor}
\usepackage{multirow}
\usepackage{threeparttable}
\def\BibTeX{{\rm B\kern-.05em{\sc i\kern-.025em b}\kern-.08em
    T\kern-.1667em\lower.7ex\hbox{E}\kern-.125emX}}
\begin{document}

\title{Cross-channel Perception Learning for H\&E-to-IHC Virtual Staining}


\author{
\IEEEauthorblockN{
Hao Yang$^{1,3}$, JianYu Wu$^{1}$, Run Fang$^{1,3}$\IEEEauthorrefmark{2}, Xuelian Zhao$^{2}$, Yuan Ji$^{2}$,\\
Zhiyu Chen$^{1}$, Guibin He$^{1}$, Junceng Guo$^{1}$, Yang Liu$^{4}$, Xinhua Zeng$^{1,3}$\IEEEauthorrefmark{2}
}
\IEEEauthorblockA{
$^{1}$College of Intelligent Robotics and Advanced Manufacturing, Fudan University\\
$^{2}$Department of Pathology, Zhongshan Hospital, Fudan University\\
$^{3}$Yiwu Research Institute, Fudan University\\
$^{4}$Department of Computer Science, The University of Toronto
}
\IEEEauthorblockA{
\IEEEauthorrefmark{2}Corresponding authors. \\
\texttt{haoyang24@fudan.edu.cn, zengxh@fudan.edu.cn}
}
}

\maketitle

\begin{abstract}
With the rapid development of digital pathology, virtual staining has become a key technology in multimedia medical information systems, offering new possibilities for the analysis and diagnosis of pathological images. However, existing H\&E-to-IHC studies often overlook the cross-channel correlations between cell nuclei and cell membranes. To address this issue, we propose a novel Cross-Channel Perception Learning (CCPL) strategy. Specifically, CCPL first decomposes HER2 immunohistochemical staining into Hematoxylin and DAB staining channels, corresponding to cell nuclei and cell membranes, respectively. Using the pathology foundation model Gigapath’s Tile Encoder, CCPL extracts dual-channel features from both the generated and real images and measures cross-channel correlations between nuclei and membranes. The features of the generated and real stained images, obtained through the Tile Encoder, are also used to calculate feature distillation loss, enhancing the model’s feature extraction capabilities without increasing the inference burden. Additionally, CCPL performs statistical analysis on the focal optical density maps of both single channels to ensure consistency in staining distribution and intensity. Experimental results, based on quantitative metrics such as PSNR, SSIM, PCC, and FID, along with professional evaluations from pathologists, demonstrate that CCPL effectively preserves pathological features, generates high-quality virtual stained images, and provides robust support for automated pathological diagnosis using multimedia medical data.
\end{abstract}


\section{Introduction}
\label{sec:intro}

The integration of multimedia information systems with artificial intelligence is driving profound transformations by enhancing accuracy, optimizing workflows, and accelerating the development of cross-disciplinary applications \cite{hou2022interaction,liu2021cross,li2024decoding,liu2022few}. Virtual staining has become one of the focal points of research and application in this context. This technology enables the generation of high-quality stained pathological images directly from unstained samples or the transformation of one staining type into another, completely eliminating the reliance on traditional chemical staining processes \cite{martell2023deep,latonen2024virtual,li2024virtual}. By combining computational imaging techniques with advanced artificial intelligence algorithms \cite{liu2020weakly,liu2023learning,ju2023novel}, virtual staining significantly reduces diagnostic time while being more environmentally friendly. It reshapes pathological diagnostic workflows and lays an important foundation for the development of reliable and rapid multimedia medical diagnostic systems.

H\&E-to-IHC transformation is a critical task in virtual staining\cite{bai2023deep}. Hematoxylin and Eosin (H\&E) staining highlights structures such as cell nuclei and cytoplasm, while Immunohistochemistry (IHC) detects tumor-related specific proteins through antigen-antibody reactions. For example, HER2 is a tumor-associated transmembrane protein, and the accurate evaluation of HER2-positive cancers relies on precise analysis of membrane staining intensity and the proportion of positive cells \cite{ahn2020her2}. Although existing methods have made significant progress, they primarily focus on single-channel features from either cell nuclei or cell membranes, failing to effectively model the spatial distribution, morphological characteristics, and staining intensity relationships between the two\cite{li2024exploiting,chen2024pathological,ahn2020her2}. This limitation results in generated images lacking cross-channel semantic consistency. Additionally, existing models are often small-scale and lack the capacity to capture complex pathological features. Treating virtual staining as a downstream task for large foundational pathology models offers a promising solution. However, practical application scenarios, such as intraoperative frozen sections, require models to be lightweight and capable of rapid inference\cite{ma2023efficient}. Overcoming these challenges is crucial for clinical applicability of virtual staining methods.

To address these challenges, we propose a novel virtual staining approach, Cross-channel Perception Learning (CCPL). First, we introduce a dual-channel perception to separate hematoxylin channel and diaminobenzidine (DAB) channel staining information from HER2 images, enabling targeted extraction of statistical features from Focal Optical Density (FOD) maps\cite{chen2024pathological} for the cell nucleus and cell membrane. Second, we utilize a pretrained tile encoder of foundation model Gigapath\cite{xu2024whole} to distill and align output image features, significantly enhancing the model’s capacity to extract pathological semantics. Most importantly, we propose a nucleus-membrane Cross-Channel Correlation to capture the spatial distribution, morphological relationships, and staining intensity associations between the cell nucleus and cell membrane. This further ensures cross-channel pathological semantic consistency between generated and real images. Quantitative metrics and human evaluation results\cite{yang2024stephanie} demonstrate that CCPL effectively improves the quality of virtual stained images, providing robust support for automated pathological diagnosis in multimedia medical data applications. The primary contributions of this paper are summarized as follows:

\begin{itemize}
    \item We are the first to highlight the cross-channel correlation between nuclei and membranes and propose a novel CCPL strategy, which enhances the pathological semantic consistency.
    \item We incorporate the tile encoder of pathology foundation model Gigapath, into the virtual staining task for feature distillation, improving the feature extraction capabilities while ensuring lightweight and efficient inference.
    \item CCPL achieves state-of-the-art performance on quantitative metrics and human evlautions, on the BCI and MIST datasets.
\end{itemize}

\section{Related Work}
Virtual staining is a key research area in multimedia medical information systems, aiming to generate stained pathological images through computational models, eliminating the need for traditional physical staining. It can be categorized into Label-free virtual staining and Stain-to-stain transformation \cite{martell2023deep,liu2024generalized,latonen2024virtual}. Label-free virtual staining uses imaging techniques and generative models to produce stained images directly from unstained tissue samples, offering a non-destructive and eco-friendly solution for rapid tissue analysis \cite{jin2024deepdof,martell2023deep,liu2022instant}. Stain-to-stain transformation focuses on converting one staining type into another, significantly reducing reagent consumption and simplifying laboratory workflows.

Stain-to-stain transformation has advanced significantly with the development of deep learning, particularly through Generative Adversarial Networks (GANs) \cite{alajaji2024generative,liu2023amp,ma2024dsff}. Since tissue samples cannot be restained, aligned consecutive sections are often used as supervision to learn mappings between staining domains. Among stain-to-stain transformation tasks, H\&E-to-IHC transformation is particularly important. H\&E staining, which uses hematoxylin and eosin, is the standard method for observing tissue structure, while IHC staining detects biomarkers such as HER2 through antigen-antibody reactions. H\&E-to-IHC transformation allows pathologists to obtain both staining types without increasing experimental costs \cite{shen2023staindiff}. Pix2pix \cite{salehi2020pix2pix} is a baseline model, and Pyramidpix2pix \cite{liu2022bci} introduces a pyramid structure to improve image quality and multi-scale consistency. ASP \cite{li2023adaptive} proposed the Adaptive Supervised PatchNCE method to address input-target inconsistencies in H\&E-to-IHC translation. PSPStain \cite{chen2024pathological} employs a protein-aware strategy and prototype consistency learning to effectively model tumor cell membrane distribution and intensity.

\section{Methods}
\begin{figure*}[ht!]
\centering
\includegraphics[width=\textwidth]{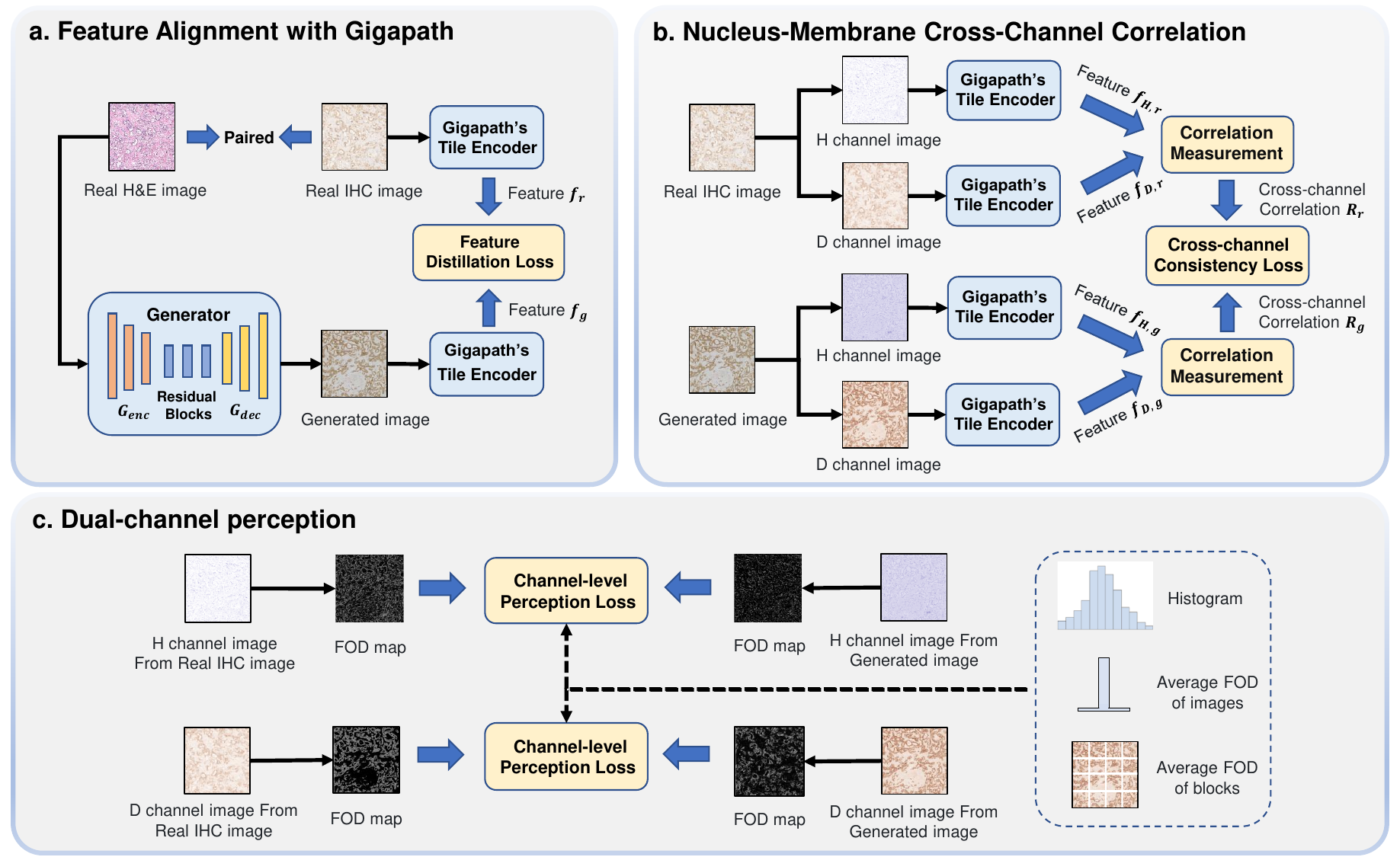}
\caption{Overview of the Cross-channel Perception. (a) Real and generated images are passed through Gigapath’s tile encoder to extract features, with the feature distillation loss calculated based on these features. (b) The IHC image is decomposed into H-channel and D-channel images, and cross-channel consistency loss is derived by evaluating the correlation between the two channels. (c) FOD maps are generated for both H and D channels, and channel-level perception loss is computed using their statistical characteristics. }
\label{3-1}
\end{figure*}

Our CCPL framework comprises three strategies: Dual-Channel Perception (DCP), Feature Distillation with Gigapath (FD), and Nucleus-Membrane Cross-channel Correlation (NMCC), as illustrated in Fig.~\ref{3-1}.

\subsection{Dual-Channel Perception}

\subsubsection{Nucleus-Membrane Staining Separation}  
To enhance the virtual staining model's ability to accurately reflect HER2 pathological features, we propose a dual-channel perception that integrates domain-specific knowledge of HER2 scoring. In pathological images, abnormal HER2 expression on the cell membrane is typically stained brown using DAB, while the cell nucleus is stained blue-purple with hematoxylin. The internal structure of the nucleus often shows uneven staining. HER2 scoring emphasizes the proportion of invasive cancer cells and the intensity of membrane staining.

We first employ traditional color deconvolution methods to separate the nucleus and membrane staining in HER2 images. The input RGB image is converted to the Hematoxylin-Eosin-DAB (HED) color space, which represents the staining intensity of hematoxylin, eosin, and DAB dyes. To isolate the respective staining components, we set the DAB and Eosin channels in the HED image to zero, leaving only the hematoxylin channel. This results in an image containing only the nucleus staining, which is then converted back to an RGB format. Similarly, an RGB image representing only the DAB channel is extracted, which corresponds to the membrane staining.

In most samples, the tumor region in the DAB channel is relatively small compared to the non-tumor regions, making it challenging to capture the tumor signal. To address this issue, we adopt Focal Optical Density (FOD)\cite{chen2024pathological} to emphasize the tumor region. For the hematoxylin channel, which represents the nucleus, the stained area is typically smaller, and the staining intensity does not vary significantly. We believe that FOD can also facilitate learning pathological information from the hematoxylin channel by enhancing the contrast of this region.

We first convert the RGB images of both channels into grayscale and then apply the following formula to convert optical density (OD) to FOD:
\begin{align}
FOD_C = 
\begin{cases} 
        (OD_C)^{\alpha_C}, & \text{if } FOD_C > T_C, \\
        0, & \text{otherwise},
    \end{cases}
\end{align}
where \( OD_C \) represents the optical density of staining channel \( C \in \{hematoxylin(H), DAB(D)\} \), and \( T_C \) (with \( T_H \) and \( T_D \)) is a global threshold applied to each channel to remove background noise. The alpha coefficients \( \alpha_H \) and \( \alpha_D \) control the degree of focusing on optical density values to enhance the significance of the cell membrane staining in the tumor region and nucleus staining.

\begin{figure*}[t]
\centering
\includegraphics[width=.9\textwidth]{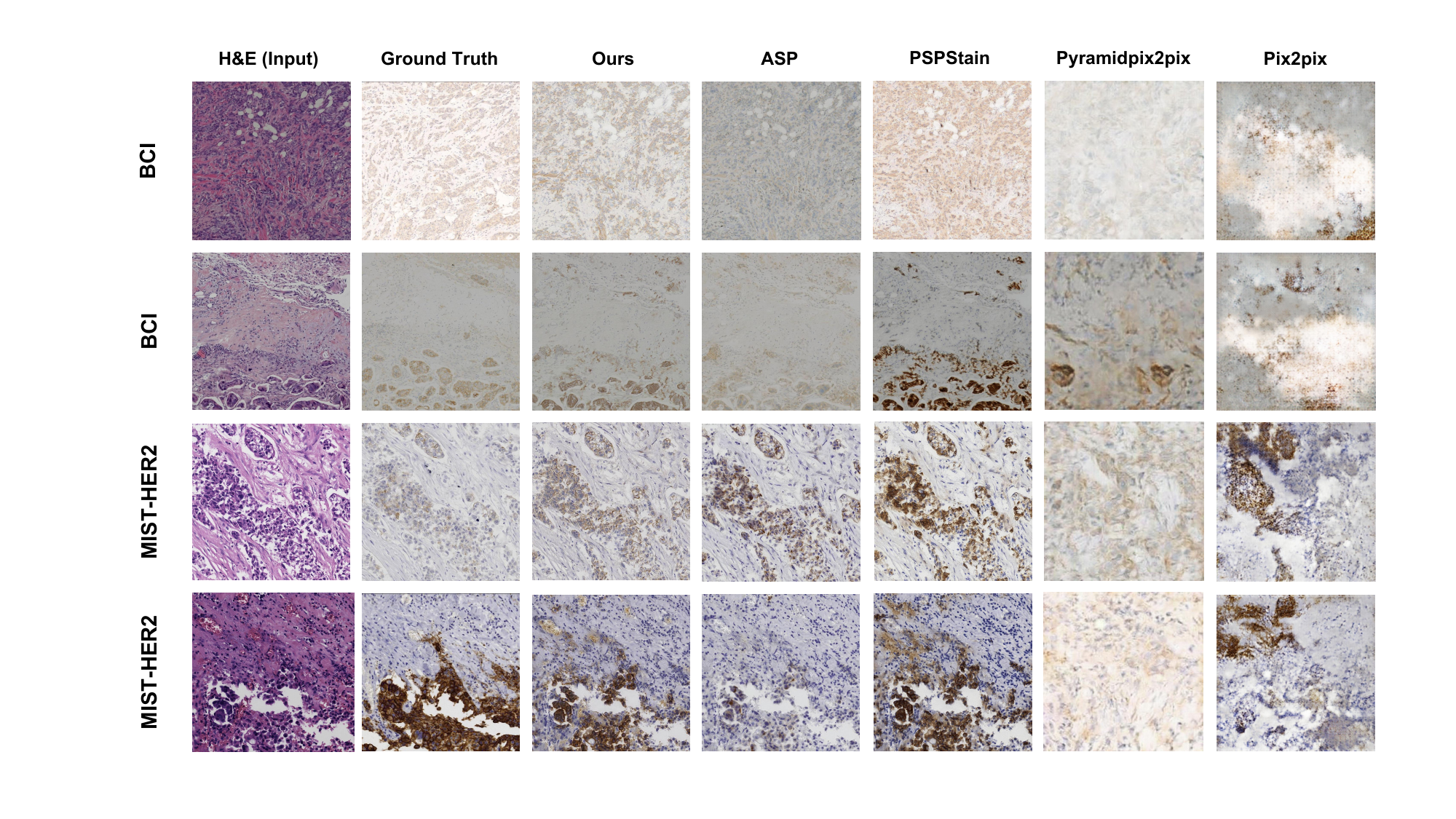}
\caption{Illustration of virtually stained results of the CCPL, ASP \cite{li2023adaptive}, PSPStain \cite{chen2024pathological}, Pyramidpix2pix \cite{liu2022bci} and Pix2Pix \cite{salehi2020pix2pix} on selected samples form the BCI \cite{liu2022bci} and MIST-HER2 \cite{li2023adaptive} datasets.}
\label{4-1}
\end{figure*}

\subsubsection{Dual-Channel Perception Loss}  
To ensure that the generated images match the real images in terms of cell nucleus and membrane staining intensity and distribution, this paper proposes a dual-channel perception loss. This loss function aligns statistical features of both the nucleus and membrane, including global optical density, local block-level optical density, and optical density distribution histograms, further ensuring consistency at multiple levels between the generated and real images.

We first compute the channel-level perception loss \(L_C\) for each channel \(C \in \{\text{H}, \text{D}\}\). The loss function \(L_C\) for channel \(C\) is expressed as:
\begin{align}
L_C = & \, \| FOD_{g,C,avg} - FOD_{r,C,avg} \|_2^2 \nonumber \\
& + \frac{1}{B} \sum_{j=1}^{B} \| FOD_{g,C,block}^{(j)} - FOD_{r,C,block}^{(j)} \|_2^2 \nonumber \\
& + \frac{1}{N} \sum_{i=1}^{N} \| H_g^{(i)} - H_r^{(i)} \|_2^2,
\end{align}
where \( FOD_{g,C,avg} \) and \( FOD_{r,C,avg} \) demote the average focal optical density values for the generated and real images in channel \( C \), respectively. \( FOD_{g,C,block}^{(j)} \) and \( FOD_{r,C,block}^{(j)} \) are the average focal optical density values for the \( j \)-th local block in channel \( C \), and \( H_g^{(i)} \) and \( H_r^{(i)} \) are the cumulative focal optical density values in the \( i \)-th histogram bin of the generated and real images. \( B \) is the number of local blocks, and \( N \) is the number of histogram bins.

To optimize the consistency of both the hematoxylin and DAB channels, we combine the losses for each channel to obtain the final dual-channel perception loss \(L_{\text{dual}}\):
\begin{align}
L_{\text{dual}} = \alpha L_{H}  + (1-\alpha) L_{D},
\end{align}
where \( \alpha \) balances the contributions of the two losses.

\subsection{Feature Alignment with Gigapath}

To leverage the strong feature extraction ability of pathological base models while keeping the model parameters at a manageable size and ensuring high inference speed, we use the pre-trained Gigapath model’s Tile Encoder. We first input the generated and real HER2 images into Gigapath’s Tile Encoder to extract their corresponding feature representations. The feature of the generated image is denoted as \( f_g \), and the feature of the real image is denoted as \( f_r \). The features extracted by Gigapath effectively capture semantic information from block-level images.

Next, we align the features of the generated and real images through feature distillation, which helps maintain the semantic consistency of virtual staining images. We use cosine similarity loss and L2 loss to align \( f_g \) and \( f_r \), with the loss functions as follows:
\begin{align}
L_{\text{fd\_cos}} &= 1 - \cos(f_g, f_r),\\
L_{\text{fd\_L2}} &= \| f_g - f_r \|_2^2.
\end{align}
The final feature distillation loss is the weighted sum of these two losses:
\begin{align}
L_{\text{fd}} = \beta L_{\text{fd\_cos}} + (1-\beta) L_{\text{fd\_L2}},
\end{align}
where \( \beta \) balances the contributions of the two losses.

\subsection{Nucleus-Membrane Cross-Channel Correlation}  

In real tissues, the cell membrane always surrounds the cell nucleus, and their distribution and morphology mutually influence each other, forming an organic whole. The HER2 expression intensity and distribution in the tumor region of the DAB channel are closely related to the morphological distribution and staining intensity of the nucleus in the H channel. Therefore, introducing cross-channel correlations in the virtual staining task is crucial for ensuring spatial alignment and relationships such as staining intensity between the cell nucleus and cell membrane channels. This ensures that the relationship between the nucleus and membrane in the generated image is consistent with that in real tissue.

In the dual-channel perception, the first step of nucleus-membrane staining separation generates the hematoxylin and DAB channel RGB images, which are input into Gigapath’s feature extractor. This yields the features for the generated image’s hematoxylin and DAB channels, denoted as \( f_{H,g} \) and \( f_{D,g} \), and the features for the real image’s hematoxylin and DAB channels, denoted as \( f_{H,r} \) and \( f_{D,r} \). We then use L2 loss and cosine similarity to measure the cross-channel correlation between the DAB and H channels in the generated and real images, denoted as \( R_{\text{g}} \) and \( R_{\text{r}} \), respectively:
\begin{align}
R_{\text{g}} = & \gamma \| f_{H,g} - f_{D,g} \|_2^2 + \nonumber \\
& (1-\gamma) (1 - \cos(f_{H,g}, f_{D,g})),\\
R_{\text{r}} = & \gamma \| f_{H,r} - f_{D,r} \|_2^2 + \nonumber \\
& (1-\gamma) (1 - \cos(f_{H,r}, f_{D,r})).
\end{align}

We then compute the L2 loss and cosine similarity between the cross-channel correlations of the generated and real images, obtaining the cross-channel consistency loss:
\begin{align}
L_{\text{cross}} = & \theta \| R_{\text{g}} - R_{\text{r}} \|_2^2 + \nonumber \\
& (1-\theta) (1 - \cos(R_{\text{g}}, R_{\text{r}})).
\end{align}

\subsection{Overall Optimization Objective}  
The final optimization objective function is as follows:
\begin{align}
L_{\text{total}} = L_{\text{adv}} + \lambda_{d} L_{\text{dual}} + \lambda_{f} L_{\text{fd}} + \lambda_{c} L_{\text{cross}} + \nonumber\\ L_{NCE} + \lambda_{m} L_{\text{SSIM}} + \lambda_{g} L_{\text{GP}}
\end{align}
where \( L_{\text{adv}} \) is the adversarial loss, and \( L_{NCE} \), \( L_{\text{SSIM}} \) represent the Noise-Contrastive Estimation loss and Structural Similarity Index Measure loss, respectively, which contribute to the enhancement of image quality. In contrast, \( L_{\text{dual}} \), \( L_{\text{fd}} \), and \( L_{\text{cross}} \) focus on maintaining pathological feature consistency.

\section{Experiments}
\subsection{Datasets and Evalution Metrics}
The experiments were conducted on two benchmark datasets: Breast Cancer Immunohistochemical (BCI) challenge \cite{liu2022bci} and Multi-IHC Stain Translation (MIST) \cite{li2023adaptive}. The BCI dataset comprises 3,980 paired H\&E-HER2 images in the training set and 995 paired images in the test set. From the MIST dataset, the MIST-HER2 subset was utilized, containing 4,643 paired samples for training and 1,001 paired samples for testing.

For model evaluation, we used Peak Signal-to-Noise Ratio (PSNR), Structural Similarity Index (SSIM), Pearson’s Correlation Coefficient (PCC), and Frechet Inception Distance (FID) as quantitative metrics, where the feature extractor for the FID metric is a pre-trained Inception network\cite{heusel2017gans}.

In addition, we invited three pathologists with extensive clinical experience to conduct a human evaluation of the clinical applicability of the generated images. Based on the HER2 scoring guidelines, we established two evaluation criteria: Positive Cell Proportion Accuracy (PCPA) and Staining Intensity Accuracy (SIA). Specifically, PCPA measures the consistency between the proportion of positive cell nuclei in the generated images and the ground truth, while SIA evaluates the consistency of staining intensity. Each pathologist scored independently, and images with significant scoring discrepancies were excluded. The final scores were averaged across the remaining images. Both metrics were rated on a 1–5 scale as follows:
\begin{itemize}
    \item 5: Almost perfectly consistent.
    \item 4: Highly consistent.
    \item 3: Generally consistent.
    \item 2: Partially consistent.
    \item 1: Completely inconsistent.
\end{itemize}

\begin{table}[t]
    \centering
    \caption{Results on BCI dataset.}\label{BCI}
    \setlength{\tabcolsep}{4pt}
    \resizebox{\linewidth}{!}{
    \begin{tabular}{l c c c c c c}
    \toprule
    \multirow{2}{*}{\textbf{Method}} & \multicolumn{4}{c}{\textbf{Quantitative metrics}} & \multicolumn{2}{c}{\textbf{Human Evaluation}} \\
    \cmidrule(lr){2-5}\cmidrule(lr){6-7}
     & SSIM$\uparrow$ & PCC$\uparrow$ & PSNR$\uparrow$ & FID$\downarrow$ & PCPA$\uparrow$ & SIA$\uparrow$ \\
    \midrule
    Pix2pix \cite{salehi2020pix2pix} & 0.2887 & 0.1260 & 14.50 & 296.20 & 1.87 & 1.96 \\
    Pyramidpix2pix \cite{liu2022bci} & 0.3631 & 0.1283 & 18.34 & 161.39 & 2.03 & 2.37 \\
    ASP \cite{li2023adaptive} & \underline{0.5025} & \textbf{0.2609} & 17.87 & \underline{54.72} & \underline{3.47} & 2.60 \\
    PSPStain \cite{chen2024pathological} & 0.4600 & 0.2187 & \underline{18.65} & \textbf{48.42} & 3.27 & \underline{3.02} \\
    Ours & \textbf{0.5280} & \underline{0.2441} & \textbf{20.42} & 57.81 & \textbf{3.68} & \textbf{3.51} \\
    \bottomrule
    \end{tabular}}
\end{table}

\subsection{Implementation Details}
CUT~\cite{park2020contrastive} is selected as the baseline model. The generator is ResNet-6Blocks, and the discriminator is PatchGAN. We trained our networks with random $512 \times 512$ crops and a batch size of four, using the Adam optimizer with a learning rate of $1\times10^{-4}$. The maximum number of training epochs was set to 80. The weight values of $\lambda_{d}$, $\lambda_{f}$, $\lambda_{c}$, $\lambda_{m}$, and $\lambda_{g}$ are 1, 1, 1, 0.05, and 10, respectively. In DCP module, the thresholds $T_H$ and $T_D$ for FOD are 0.15, $\alpha_H$ and $\alpha_D$ are 1.8, and the weight coefficient $\alpha$ for the H channel is set to 0.1.
\subsection{Comparison with Existing Methods}

As shown in Tables~\ref{BCI} and \ref{MIST}. The quantitative comparisons with existing methods on bechmark datasets highlights the effectiveness and superiority of the proposed CCPL. Specifically, it achieves the highest scores in terms of both PSNR and SSIM metrics, indicating that the generated images have lower pixel-level errors and better structural fidelity. On the PCC metric, CCPL ranks first and second on the BCI and MIST-HER2 datasets, respectively, demonstrating its effectiveness in capturing the linear correlations between the generated and ground-truth images. Furthermore, CCPL achieves the lowermost FID scores on the BCI dataset, reflecting its ability to produce pathological feature distributions closely aligned with real data. Human evaluation further validates that CCPL generates results highly similar to the ground truth in critical diagnostic pathological features. 

Furthermore, we show the staining results of CCPL and existing mainstream methods on selected samples from BCI \cite{liu2022bci} and MIST-HER2 \cite{li2023adaptive} datasets in Fig.~\ref{4-1} to qualitatively assess its effectiveness and application prospects. Among them, columns 1 and 2 are the original H\&E images and the groud truth of staining, respectively, while columns 3-7 represent the staining results of the proposed CCPL, ASP \cite{li2023adaptive}, Pix2pix \cite{salehi2020pix2pix},  PSPStain \cite{chen2024pathological}, and Pyramidpix2pix \cite{liu2022bci}, respectively. CCPL achieved gripping visualization on all four samples, and was closer to GT in terms of staining intensity and proportion of positive cells.

\begin{table}[t]
\centering
\caption{Results on MIST-HER2 dataset.}\label{MIST}
\setlength{\tabcolsep}{4pt}
\resizebox{\linewidth}{!}{
\begin{tabular}{l c c c c c c}
\toprule
\multirow{2}{*}{\textbf{Method}} & \multicolumn{4}{c}{\textbf{Quantitative metrics}} & \multicolumn{2}{c}{\textbf{Human Evaluation}} \\
\cmidrule(lr){2-5}\cmidrule(lr){6-7}
 & SSIM$\uparrow$ & PCC$\uparrow$ & PSNR$\uparrow$ & FID$\downarrow$ & PCPA$\uparrow$ & SIA$\uparrow$ \\
\midrule
Pix2pix \cite{salehi2020pix2pix}  & 0.0788 & 0.0604 & 12.47 & 251.75 & 2.40 & 2.33 \\
Pyramidpix2pix \cite{liu2022bci} & 0.1677 & 0.0042 & 13.62 & 255.14 & 2.23 & 2.47 \\
ASP \cite{li2023adaptive} & \underline{0.1971} & 0.1231 & 14.14 & 50.13 & 3.23 & 3.10 \\
PSPStain \cite{chen2024pathological} & 0.1874 & \underline{0.1352} & \underline{14.18} & \underline{46.47} & \underline{3.37} & \underline{3.33} \\
Ours & \textbf{0.1998} & \textbf{0.1388} & \textbf{14.70} & \textbf{46.15} & \textbf{3.72} & \textbf{3.63} \\
\bottomrule
\end{tabular}}
\end{table}

\begin{table}[t]
\centering
\caption{Results of Ablation Study.}\label{ablation}
\setlength{\tabcolsep}{2.5pt}
\resizebox{\linewidth}{!}{
\begin{tabular}{l c c c c c c c c c}
\toprule
\multirow{2}{*}{\textbf{ID}} & \multicolumn{3}{c}{\textbf{Components}} & \multicolumn{4}{c}{\textbf{Quantitative Metrics}} & \multicolumn{2}{c}{\textbf{Human Evaluation}} \\
\cmidrule(lr){2-4}\cmidrule(lr){5-8}\cmidrule(lr){9-10}
 & FD & DCP & NMCC & SSIM$\uparrow$ & PCC$\uparrow$ & PSNR$\uparrow$ & FID$\downarrow$ & PCPA$\uparrow$ & SIA$\uparrow$ \\
\midrule
1 & \ding{55} & \ding{55} & \ding{55} & 0.5045 & 0.2166 & 19.96 & 60.59 & 2.21 & 2.13 \\
2 & $\checkmark$ & \ding{55} & \ding{55} & 0.4947 & 0.2256 & 20.68 & 40.05 & 2.76 & 2.74 \\
3 & $\checkmark$ & \ding{55} & $\checkmark$ & 0.5383 & 0.2327 & 19.70 & 63.83 & 3.23 & 2.97 \\
4 & $\checkmark$ & $\checkmark$ & $\checkmark$ & 0.5280 & 0.2441 & 20.42 & 57.81 & 3.68 & 3.51 \\
\bottomrule
\end{tabular}}
\end{table}

\subsection{Ablation Study}

To validate the effectiveness of the components in the CCPL framework, we conducted an ablation study by adding or removing FD, DCP, and NMCC modules, and evaluated their impact on image quality, as shown in table \ref{ablation}. The results indicate that FD significantly improves the FID score, reflecting better semantic consistency (Model 1 vs. Model 2). NMCC enhances SSIM and PCC scores by modeling cross-channel correlations. Although FID scores slightly decrease in this case, human evaluation suggests that NMCC positively contributes to preserving pathological features. We attribute the FID decline to its reliance on the Inception-based feature extractor, which is not well-suited for extracting pathology-specific features. DCP further improves PSNR and PCC scores by enhancing single-channel detail representation (Mode 3 vs. Model 4). The complete model achieves balanced improvements across all metrics, validating the synergistic effects of these components in enhancing image quality and maintaining pathological consistency for virtual staining tasks.

\section{Conclusion}
This study proposes a novel virtual staining method that leverages Gigapath’s Tile Encoder to introduce a loss function based on feature differences between generated and real stained images, while maintaining high efficiency in the inference process. The method decomposes HER2 immunohistochemical staining into Hematoxylin and DAB staining channels corresponding to cell nuclei and cell membranes, respectively, extracts feature distances between the channels, and learns cross-channel correlations between nuclei and membranes. Additionally, statistical analysis of the focal optical density maps for both channels is conducted to ensure consistency in staining distribution and intensity. Experimental results demonstrate that this method effectively preserves pathological features and generates high-quality virtual stained images, as evidenced by quantitative metrics such as PSNR, SSIM, and FID, as well as professional evaluations by pathologists. This achievement provides robust support for automated pathological diagnosis based on multimedia medical data and further enhances the clinical applicability of virtual staining. Our future work will focus on improving the representation and learning of cross-channel correlations in models. Furthermore, incorporating diagnostic texts related to pathological features generated by foundational pathology models as auxiliary information for virtual staining is another promising direction we aim to explore.

\bibliographystyle{IEEEbib}
\bibliography{ref}

\end{document}